\documentclass[letterpaper, 10 pt, conference]{ieeeconf}
\IEEEoverridecommandlockouts
\pdfoutput=1
\usepackage{xparse} 
\usepackage{amsmath} 

\providecommand{\norm}[1]{\lVert#1\rVert}
\usepackage{amssymb}
\usepackage{amsfonts} 
\usepackage{pgfplots}
\usepackage{bbm}
\usepackage{color}
\usepackage{tikz}
\usetikzlibrary{plotmarks}
\usetikzlibrary{bayesnet}

\usepackage{graphicx}

\usepackage{subcaption}
\usepackage[noadjust]{cite}

\usepackage{url}
\usepackage{framed}

\usepackage{algpseudocode}
\usepackage{algorithm}

\usepackage{threeparttable}
\usepackage{booktabs}
\usepackage{arydshln}

\usepackage{balance}
\usepackage{enumerate}   

\usepackage{pdfpages}

\definecolor{pyplotred}{RGB}{255, 127, 14}
\definecolor{pyplotblue}{RGB}{31, 119, 180	}
\definecolor{pyplotgreen}{RGB}{44, 160, 44	}
\definecolor{mygray}{rgb}{0.4,0.4,0.4}

\title{\bf Fast Manipulability Maximization Using\\ Continuous-Time Trajectory Optimization}
\author{Filip Mari\'c$^{* \dagger}$, Oliver Limoyo$^*$, Luka Petrovi\'c$^{\dagger}$, Trevor Ablett$^*$, Ivan Petrovi\'c$^{\dagger}$, and Jonathan Kelly$^*$
  \thanks{This research was supported in part by a Dean's Catalyst Professorship from the University of Toronto and the European Regional Development Fund under the grant KK.01.1.1.01.0009 (DATACROSS).}
  \thanks{ $^*$ Filip Mari\'c, Oliver Limoyo, Trevor Ablett, and Jonathan Kelly are with the University of Toronto, Institute for Aerospace Studies, Space and Terrestrial Autonomous Robotic Systems Laboratory, Canada. \{\texttt{<first name>.<last name>@robotics.utias.utoronto.ca}\}}
  \thanks{ $^\dagger$ Filip Mari\'c, Luka Petrovi\'c, and Ivan Petrovi\'c are with the University of Zagreb, Faculty of Electrical Engineering and Computing, Laboratory for Autonomous Systems and Mobile Robotics, Croatia. \{\texttt{<first name>.<last name>@fer.hr}\}}
}
\begin{document}

\maketitle
\thispagestyle{empty}
\pagestyle{empty}

\begin{abstract}
  A significant challenge in manipulation motion planning is to ensure agility in the face of unpredictable changes during task execution.
  This requires the identification and possible modification of suitable joint-space trajectories, since the joint velocities required to achieve a specific end-effector motion vary with manipulator configuration.
  For a given manipulator configuration, the joint space-to-task space velocity mapping is characterized by a quantity known as the manipulability index.
  In contrast to previous control-based approaches, we examine the maximization of manipulability during planning as a way of achieving adaptable and safe joint space-to-task space motion mappings in various scenarios.
  By representing the manipulator trajectory as a continuous-time Gaussian process (GP), we are able to leverage recent advances in trajectory optimization to maximize the manipulability index during trajectory generation.
  Moreover, the sparsity of our chosen representation reduces the typically large computational cost associated with maximizing manipulability when additional constraints exist.
  Results from simulation studies and experiments with a real manipulator demonstrate increases in manipulability, while maintaining smooth trajectories with more dexterous (and therefore more agile) arm configurations.
\end{abstract}

\section{Introduction}\label{section:introduction}

Motion planning is a fundamental challenge for robotic manipulators executing complex tasks.
To perform a task successfully, the motion planner must generate a joint space trajectory that respects constraints induced by the task (e.g., collision avoidance).
The existence of these constraints may cause the planning algorithm to generate trajectories that contain configurations with suboptimal joint space-to-task space mappings.
Consequently, in scenarios where the manipulator is operating autonomously in non-static environments (e.g., during collaborative task execution), large joint motions may be required in order for the manipulator to adapt to unexpected changes in constraints during task execution.
A configuration's capacity for movement in the task space can be inferred from the manipulability ellipsoid~\cite{sciavicco2012modelling}, whose axis lengths give a measure of how effectively joint velocities map to directions in task space.
The manipulability index introduced by Yoshikawa in~\cite{yoshikawa1985manipulability} is proportional to this ellipsoid's volume and is commonly used in pose-tracking controllers to avoid particularly unfavourable mappings that are known as \textit{singularities}.
It follows that, by maximizing manipulability, we can ensure that the manipulator possesses a higher level of overall dexterity throughout the planned trajectory---while avoiding large and potentially hazardous joint movements.
This enables rapid and predictable manipulator responses in safety-critical applications as diverse as robotically-assisted surgery~\cite{san2007mechanical} or satellite capture~\cite{nanos2015avoiding}.
Rather than relying on the incorporation of singularity avoidance in the tracking controller itself, our method generates a motion plan that preemptively maximizes the overall manipulability throughout the manipulator's trajectory.

Our goal is to optimize the trajectory such that, in the face of unexpected task changes (e.g., to the final end-effector pose), the manipulator is able to adapt its motion with minimal joint position changes.
\begin{figure}
  \centering
  \includegraphics[trim={0 3cm 0 3cm},clip, width=\columnwidth]{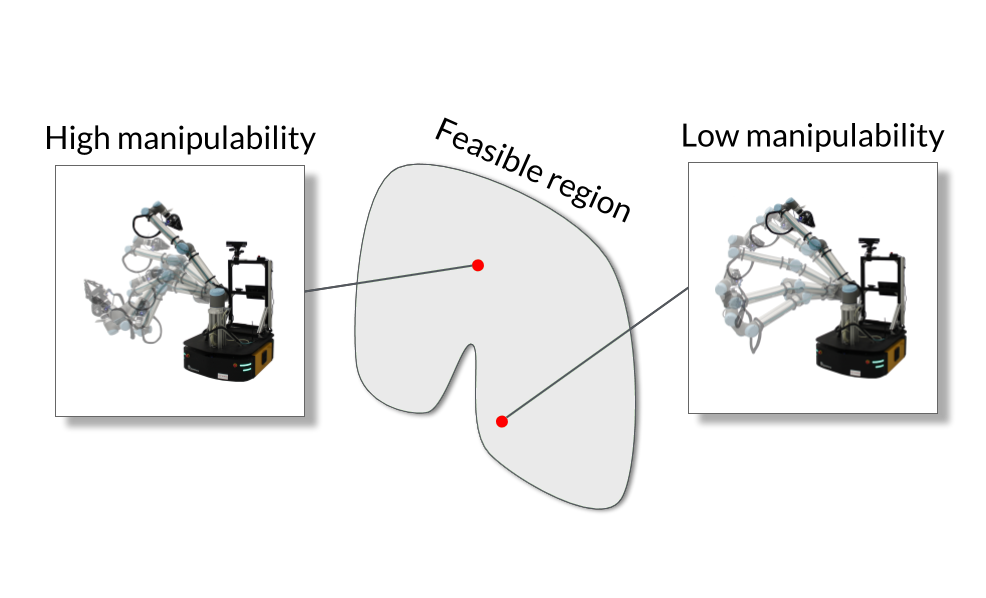}
  \caption{Comparison of two solutions for reaching a position goal from a given near-singular starting configuration (caused by fully extending the arm initially).
    The right image shows a solution based purely on inverse kinematics, which maintains low manipulability throughout.
    The image on the left shows a trajectory generated by our method, which avoids excessive arm extension.}\label{fig:IRL}
  \vspace{-5mm}
\end{figure}
The arm trajectory shown on the right side of Fig.~\ref{fig:IRL} is an example in which the configuration is initially (and throughout the motion) nearly singular, with the arm fully extended. Hence, even small movements of the end-effector along the extended axis will result in large joint velocities at the elbow.
On the left side of Fig.~\ref{fig:IRL} is a trajectory with high manipulability that reaches the same end-effector goal position in task space while avoiding the elbow singularity.
By choosing to represent the complete joint trajectory as a sample from a continuous-time Gaussian process (GP) \cite{rasmussen2004gaussian}, the manipulability maximization problem defined above can be formulated as probabilistic inference; a \textit{maximum a posteriori} (MAP) estimator can be used~\cite{barfoot2014batch, gpmp-ijrr} to find a solution that is, locally, relatively far from near-singular regions while also enforcing a notion of smoothness through a trajectory prior.
We build on the approach presented in~\cite{gpmp-ijrr}, by introducing a likelihood factor which helps avoid low manipulability configurations induced by task constraints.
To the best of the authors' knowledge, this is the first method to directly integrate manipulability maximization within a trajectory optimization formulation. We make the following contributions:
\begin{enumerate}[(i)]
  \itemsep2pt
\item we formulate manipulability maximization as a continuous-time trajectory optimization problem,
\item we demonstrate that this approach can be applied to a  variety of planning scenarios,
\item we show that, for our chosen trajectory representation, the problem can be efficiently solved, and
\item we compare our approach to existing singularity avoidance and manipulability maximization techniques.
\end{enumerate}

\section{Related Work}\label{section:related_work}
Manipulability maximization has been extensively studied from the perspective of robust kinematic control.
Redundancy resolution schemes such as~\cite{chiaverini1997singularity} and \cite{marani2002real} have long been used for singularity avoidance, and consequently to increase overall manipulability.
More recently, quadratic programming (QP) has been examined as an efficient method for manipulability maximization in constrained inverse kinematics solvers~\cite{dufour2017integrating, zhang2016qp, jin2017manipulability}.
These kinematic control policies are efficient for end-effector tracking tasks, although their success ultimately depends on the trajectory and on manipulator redundancy.
Moreover, many common tasks cannot be efficiently defined in this manner (e.g., the task of reaching a goal configuration while avoiding an obstacle).

Previous attempts to generate joint-space trajectories or paths with high manipulability have resorted to methods that suffer from high computational cost. 
For example, in~\cite{menasri2013path}, a maximum manipulability discrete joint-space path for a five degrees-of-freedom (DOF) manipulator is produced by a genetic algorithm.
The resolution-complete search in~\cite{guilamo2006manipulability} generates a sequence of high manipulability configurations from an end-effector path in obstacle-free environments.
In~\cite{rybus2013experimental}, a singularity-free joint-space path is generated by parameterizing the end-effector trajectory using Bezier curves and finding optimal configurations through simulated annealing. 
In~\cite{osa2017guiding} a manipulability cost is learned as one of several cost terms in a formulation similar to CHOMP~\cite{ratliff2009chomp}, with the aim of completing a `disentangling' task from a provided demonstration.
Additional interpolation may also be required in order to ensure smooth transitions between the states generated by these methods.
However, no prior information will be available on the manipulability or collision of the arm with the environment for these interpolated states.

Trajectory optimization algorithms~\cite{ratliff2009chomp,kalakrishnan2011stomp,park2012itomp,schulman2013finding} minimize a cost function composed of both optimality and feasibility terms and have been used for online planning and replanning due to their low computational demands.
However, avoiding singularities presents a difficult problem for such approaches, since they require dense discretizations to produce smooth and feasible solutions when handling complicated constraints.
This is especially true when the problem is additionally constrained by the end-effector pose or by the need to avoid obstacles in the environment.

A continuous-time trajectory representation can sidestep many of these difficulties by reducing the number of states used in the optimization.
Mukadam et al.~\cite{gpmp-ijrr} use a GP trajectory representation, which allows them to treat the motion planning problem as probabilistic inference on a factor graph; as such, they are able to interpolate over a trajectory and generate additional gradient information.
Consequently, highly constrained problems can be solved efficiently using a MAP estimator while requiring only a relatively small number of states.
The resulting trajectory can be queried at any point, allowing the robot's manipulability to be monitored throughout.

\section{Manipulability}
\label{section:singularities}

Consider a joint configuration $\boldsymbol{\theta}_i$ as the state of a trajectory $\boldsymbol{\theta}$ at time $t_i$.
The kinematic relationship between configuration space and task space velocities at $\boldsymbol{\theta}_i$ for an $n$-DOF robot is defined as
\vspace{-3mm}
\begin{align}
  \label{eq:jacobian}
  \dot{\mathbf{x}} = \mathbf{J}\left(\boldsymbol{\theta}_i\right)\boldsymbol{\omega} ,
\end{align}
where $\mathbf{J}\left( \boldsymbol{\theta}_i\right) \in \mathbb{R}^{p\times n}$ is the manipulator Jacobian matrix at $\boldsymbol{\theta}_i$, while $\boldsymbol{\omega} \in \mathbb{R}^{n}$ and $\dot{\textbf{x}} \in \mathbb{R}^{p}$ are the configuration and task space velocities at $t_i$, respectively.
Now, consider an $n$-dimensional sphere in the space of unit joint velocities $\|\boldsymbol{\omega}\|^2 = 1$; using Eq.~\eqref{eq:jacobian} we can define the mapping to the Cartesian (task) velocity space as
\begin{align}\label{eq:ellipsoid}
  \|\boldsymbol{\omega}\|^2 = \dot{\mathbf{x}}^T\left(\mathbf{J}\mathbf{J}^T\right)^{-1}\dot{\mathbf{x}}.
\end{align}
From Eq.\ \eqref{eq:ellipsoid}, we see that the scaling of joint velocities to the task space depends on the conditioning of the symmetric positive semi-definite matrix $\mathbf{J}\mathbf{J}^T$.
Manipulability provides a computationally tractable measure of the conditioning of $\mathbf{J}\mathbf{J}^T$ for any joint configuration~\cite{yoshikawa1985manipulability}.
\begin{figure}
  \centering
  \def\svgwidth{0.5\columnwidth}
  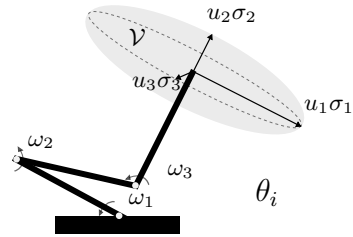
  \vspace{1mm}
  \caption{Illustration of the manipulability ellipsoid of volume $\mathcal{V}$ for a manipulator end-effector at configuration $\boldsymbol{\theta}_i$.
    Larger axis lengths indicate higher mobility.}\label{fig:mnp_ellipsoid}
  \vspace{-5mm}
\end{figure}

\subsection{Manipulability}

The matrix $\mathbf{J}\mathbf{J}^T$ in Eq.\ \eqref{eq:ellipsoid} also defines the \textit{manipulability ellipsoid}~\cite{sciavicco2012modelling} of the end-effector.
The principal axes $\sigma_1\mathbf{u}_1, \sigma_2 \mathbf{u}_2,\dots, \sigma_p\mathbf{u}_p$ of this ellipsoid can be determined through singular value decomposition of $\mathbf{J} = \mathbf{U}\boldsymbol{\Sigma}\mathbf{V}^T$.
The manipulability measure (index) of a given kinematic chain at $\boldsymbol{\theta}_i$ is defined as 
\begin{align}\label{eq:mnp}
  m = \sqrt{\det{\left(\mathbf{\mathbf{J}}\mathbf{J}^T\right)}} = \sigma_1 \sigma_2 \dots \sigma_k \dots \sigma_p,
\end{align}
and is proportional to the volume, $\mathcal{V}$, of the manipulability ellipsoid~\cite{sciavicco2012modelling}. 
Here, $\sigma_k \geq 0 $ is the $k$-th largest singular value of $\mathbf{J}$, while $\mathbf{u}_{k}$ is the $k$-th column vector of $\mathbf{U}$.
A low manipulability corresponds to a low volume of the manipulability ellipsoid, inhibiting motion in the task space. An example of the manipulability ellipsoid of the end-effector frame of a simple manipulator is depicted in Fig.\ \ref{fig:mnp_ellipsoid}. 
The gradient of Eq.\ \eqref{eq:mnp} can be calculated numerically~\cite{dufour2017integrating}, but it is also possible to derive the gradient analytically with respect to the $j$-th joint of the configuration $\boldsymbol{\theta}_i$ using Jacobi's identity~\cite{marani2002real},
\begin{align}\label{eq:mnp_jacobian}
  \frac{\partial\,m}{\partial\,\theta_{i,j}} =  m \,
  Tr\left(
  \frac{\partial \mathbf{J}}{\partial \theta_{i,j}}\mathbf{J}^{\dagger}
  \right) \ .
\end{align}
Moreover, the components of Eq.\ \eqref{eq:mnp_jacobian}, $\mathbf{J}$ and $\frac{\partial \mathbf{J}}{\partial\theta_{i,j}}$, can be calculated via geometrical methods~\cite{hourtash2005kinematic}.

\subsection{Singularities}

The concept of manipulability relates directly to the conditioning of the manipulator Jacobian matrix.
Configurations that result in the matrix $\mathbf{J}\mathbf{J}^T$ in Eq.\,\eqref{eq:ellipsoid} being non-invertible are termed \textit{singularities}. 
Consider a kinematic chain and corresponding manipulability ellipsoid with a volume $\mathcal{V} \propto m$, as shown in Fig.\ \ref{fig:mnp_ellipsoid}. If the ellipsoid contains one or more zero-length principal axes, it follows that $\mathcal{V} = 0$ and $m = 0$; configurations yielding such ellipsoids are known as \textit{singular configurations}.
We can define a minimum acceptable ellipsoid volume $\mathcal{V}_S \in \mathbb{R}_{+}$, and regard configurations that result in a manipulability $m < m\left(\mathcal{V}_S\right)$ to be \textit{nearly singular}. 

\begin{figure}
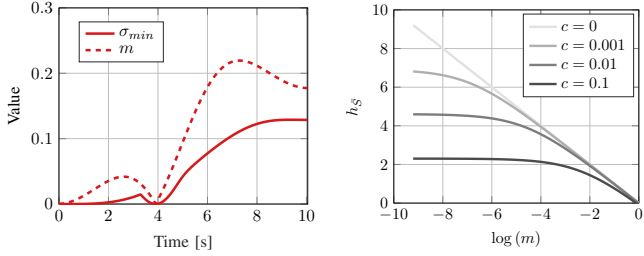

  \begin{subfigure}[]{0.49\columnwidth}
    \centering
    \resizebox {\columnwidth}{!}{\input{figures/lowest_eigenvalue.tex}}	
  \end{subfigure}
  \begin{subfigure}[]{0.49\columnwidth}
    \resizebox {\columnwidth}{!}{\input{figures/sim_mnp/final/log1c.tex}}
  \end{subfigure}
  \caption{Left: Comparison of the manipulability measure (dashed line) and the smallest singular value of the Jacobian throughout a trajectory. Right: Shape of the likelihood function in Eq.~\eqref{eq:mnp_likelihood}, which depends on the parameter $c$.}\label{fig:mnp_eig}
  \vspace{-5mm}
\end{figure}

Conversely, a high manipulability value does not guarantee that a configuration is not nearly singular, as an ellipsoid with one `degenerate' (i.e., of very small magnitude) axis may still have a large overall volume.
The volume of any manipulability ellipsoid for the chain is bounded by the value $\mathcal{V}_{max}$, determined by the chain's kinematic parameters.
Assuming that the axes $\sigma_1\textbf{u}_1, \sigma_2\textbf{u}_2,\dots, \sigma_p\textbf{u}_p$ are of an acceptable length for all such ellipsoids, we infer that configurations whose ellipsoid volume are sufficiently close to $\mathcal{V}_{max}$ are \emph{not} nearly singular (labelled as $\bar{S}$).
Fig.~\ref{fig:mnp_eig} compares the smallest singular value of the manipulator Jacobian to the manipulability measure (index) throughout a sample trajectory; the manipulability measure roughly follows the magnitude of the smallest singular value, matching its peaks and troughs.

\section{Manipulability Maximization Formulation}\label{section:trajopt}
If we consider the joint space trajectory $\boldsymbol{\theta}$ as a function which maps every time instance $0 \leq t \leq T$ to a configuration $\boldsymbol{\theta}(t)$, manipulability maximization can be formulated as trajectory optimization,
\begin{equation}
  \begin{aligned}
    \label{eq:trajopti}
    & \underset{\boldsymbol{\theta}(t)}{\text{minimize}} & & \mathcal{F}\left[\boldsymbol{\theta}(t)\right] + \lambda \mathcal{M}\left[\boldsymbol{\theta}(t)\right] + \mu \mathcal{C}\left[\boldsymbol{\theta}(t)\right],
  \end{aligned}
\end{equation}
where $\mathcal{F}\left[\boldsymbol{\theta}(t)\right]$ is a cost functional encoding smoothness, $\mathcal{M}\left[\boldsymbol{\theta}(t)\right]$ is a cost functional relating manipulability to the trajectory space, and $\mathcal{C}\left[\boldsymbol{\theta}(t)\right]$ is a cost functional that enforces collision avoidance (necessary in many environments).

\subsection{Representing the Trajectory as a GP}
\label{section:traj_rep}

Using the definition of trajectory optimization in Eq.~\eqref{eq:trajopti}, we follow the derivation in~\cite{barfoot2014batch}, representing the continuous-time trajectory as a sample from a vector-valued GP, $\boldsymbol{\theta}(t) \sim \mathcal{GP}(\boldsymbol{\mu}(t), \textbf{K}(t, t^{\prime})) $, with mean $\boldsymbol{\mu}(t)$ and covariance $ \textbf{K}(t, t^{\prime})$, generated by a linear time-varying stochastic differential equation (LTV-SDE),
\begin{equation}\label{ltvsde}
  \dot{\boldsymbol{\theta}}(t) = \textbf{A}(t) \boldsymbol{\theta}(t) + \boldsymbol{u}(t) + \textbf{F}(t) \boldsymbol{w}(t),
\end{equation}
where $\boldsymbol{A}$ and $\boldsymbol{F}$ are system matrices, $\textbf{u}$ is a known control input and $\boldsymbol{w} \sim \mathcal{N}(\mathbf{0}, \mathbf{Q}_c)$.
For any set of times $\mathbf{t}$, the corresponding \textit{support states} $\boldsymbol{\theta} = \left[\boldsymbol{\theta}_0 \dots \boldsymbol{\theta}_N\right]^{T}$ can be matched to an exponential prior distribution resulting from the system in Eq.~\eqref{ltvsde}, with the mean $\boldsymbol{\mu}$ and kernel $\textbf{K}$:
\begin{equation}
  p(\boldsymbol{\theta}) \propto \exp  \{  -\frac{1}{2} \norm{\boldsymbol{\theta} - \boldsymbol{\mu} }_\textbf{K}^2   \}.
  \label{eq:prior}
\end{equation}
The Markovian property of the process in Eq.\ \eqref{ltvsde} allows us to factor the prior in Eq.~\eqref{eq:prior} as 
\begin{align}\label{eq:prior2}
  p ( \boldsymbol { \theta } ) \propto f _ { 0 } ^ { p } \left( \boldsymbol { \theta } _ { 0 } \right) f _ { N } ^ { p } \left( \boldsymbol { \theta } _ { N } \right) \prod _ { i = 0 } ^ { N - 1 } f _ { i } ^ { g p } \left( \boldsymbol { \theta } _ { i } , \boldsymbol { \theta } _ { i + 1 } \right).
\end{align}
where $f _ { 0 } ^ { p }$ and $f _ { N } ^ { p }$ define the prior distributions on the start and end states, and $f _ { i } ^ { g p }$ is the GP prior factor as defined in~\cite{gpmp-ijrr}.
Furthermore, this property allows for interpolation of the trajectory in $\mathcal{O}(1)$ time~\cite{barfoot2014batch}.

\subsection{Manipulability Likelihood Function}\label{subsection:likelihood}

Representing a trajectory using the GP in Eq.\,~\eqref{ltvsde} allows us to intuitively (locally) search for high manipulability variants using probabilistic inference.
We formulate the likelihood of the trajectory $\boldsymbol{\theta}$ being free of singular (low manipulability) configurations, denoting this event by $\bar{S}$.
The singularity factor $f _ { i } ^ {\bar{S}}$ defines this likelihood for the support states, while $f _ {\tau}^{ \bar{S}}$ does so for the interpolated states at times $t _ { i } < \tau < t _ { i + 1 }$:
\begin{equation}\label{eq:singularity_likelihood2}
\begin{split}
f_{i}^{\bar{S}} =  &\exp{ \bigl\{   -\frac{1}{2}  \norm{ h_{\bar{S},i}(\boldsymbol { \theta } _ { i }) }_ {\Sigma_{\bar{S}}} ^ { 2 } \bigr\} }, \\
f _ { \tau } ^ {\bar{S}} = &\exp \bigl\{ - \frac { 1 } { 2 } \bigl\| h_{\bar{S},i} ( \boldsymbol { \mu } ( \tau ) + \boldsymbol { \Lambda } ( \tau ) ( \boldsymbol { \theta } _ { i } - \boldsymbol { \mu } _ { i } )  \bigr. \bigr. \\
 \bigl. \bigl. & \quad  \quad \boldsymbol { \Psi } ( \tau ) (\boldsymbol { \theta } _ { i + 1 } - \boldsymbol { \mu } _ { i + 1 }) \bigr\|_ {\Sigma_{\bar{S}}} ^ { 2 } \bigr\}
\end{split}
\vspace{-2mm}
\end{equation}

Matrices $\boldsymbol { \Lambda }$ and $\boldsymbol { \Psi }$ are defined as in~\cite{gpmp-ijrr}, and $h_{\bar{S},i}$ is the cost function.
Parametrizing the distributions of the factors in Eq.\ \eqref{eq:singularity_likelihood2} using the manipulability index allows us to find a maximum manipulability posterior using a MAP estimator. 

The manipulability index may vary by several orders of magnitude throughout a trajectory, as it is proportional to a $p$-dimensional volume. Additionally, large changes in manipulability can be caused by small shifts in the manipulator configuration.
This presents a problem when maximizing a likelihood of the form in Eq.~\eqref{eq:singularity_likelihood2}, as $\norm{ h_{\bar{S},i} }^2_{{\Sigma_{\bar{S}}}}$ needs to be minimized. 
To this end, we choose the cost $h_{\bar{S},i}$ to be logarithmic,
\vspace{-2mm}
\begin{align}\label{eq:mnp_likelihood}
  h_{\bar{S},i} = \log\left(\frac{m_{max} + c}{m + c}\right) ,
  \vspace{-2mm}
\end{align}
where the value $m_{max}$ is the manipulability value at $\mathcal{V}_{max}$ or higher.
The constant $c$ serves to limit the log-linear cost change when the value is below a certain order of magnitude, resulting in the gradient
\begin{align}\label{eq:mnp_likelihood_grad}
  \vspace{-2mm}
  \frac{\partial h_{\bar{S},i}}{\partial \theta_{i,j}} = - \frac{m}{m + c} \,
  Tr\left(
  \frac{\partial \mathbf{J}}{\partial \theta_{i,j}}\mathbf{J}^{\dagger}
  \right) \ .
  \vspace{-2mm}
\end{align}
This reduces the range of possible gradient values, simplifying the choice of weighing parameter ${\Sigma_{\bar{S}}}$.
In Fig.~\ref{fig:mnp_eig}, on the right we can see the effect of the value $c$ on the overall cost in Eq.~\eqref{eq:mnp_likelihood_grad}.

\subsection{Optimizing the Trajectory}
\label{subsection:optimization}

By representing the initial trajectory as the prior in Eq.~\eqref{eq:prior2}, we can find a continuous trajectory which (locally) maximizes the likelihood in Eq.~\eqref{eq:singularity_likelihood2} by using a MAP estimator over the support states
\vspace{-1mm}
\begin{equation}\label{eq:MAP}
  \boldsymbol{\theta}^* = \underset{\boldsymbol{\theta}}{\arg\max} \, p(\boldsymbol{\theta})\,l ( \boldsymbol { \theta } ; \bar{S}  ) \, l ( \boldsymbol { \theta } ; \bar{C}  ).
\end{equation}
\vspace{-1mm}
The maximization in Eq.\ \eqref{eq:MAP} can be achieved by computing
\begin{equation}\label{eq:MAP_log}
\vspace{-1mm}
\begin{split}
    \boldsymbol{\theta}^* = \underset{\boldsymbol{\theta}}{\arg\min}\,& \frac{1}{2} \bigl\|\boldsymbol{\theta} - \boldsymbol{\mu} \bigr\|_\textbf{K}^2 + \frac{1}{2}\bigl\|\mathbf{h}_{\bar{S}}\left(\boldsymbol{\theta}\right) \bigr\|^2_{{\boldsymbol{\Sigma}_{\bar{S}}}}\\
    &+ \frac{1}{2}\bigl\|\mathbf{h}_{\bar{C}}\left(\boldsymbol{\theta}\right) \bigr\|^2_{{\boldsymbol{\Sigma}_{\bar{C}}}}.
\end{split}
\vspace{-1mm}
\end{equation}
This is equivalent to the negative log of the posterior, where $\mathbf{h}_{\bar{S}}$ and $\mathbf{h}_{\bar{C}}$ are vectors of manipulability and collision costs for both support and interpolated states.
By linearizing Eq.~\eqref{eq:MAP_log} around the current trajectory $\boldsymbol{\theta}$, we arrive at a least squares problem.

Barfoot et.\ al~\cite{barfoot2014batch} show that the GP generated by Eq.~\eqref{ltvsde} results in a kernel $\textbf{K}$ that induces sparsity in the problem defined by Eq.~\eqref{eq:MAP_log}, making it easily solvable using sparse Cholesky decomposition.
Mukadam et.\ al~\cite{gpmp-ijrr} show that a constant velocity prior (i.e., forming a straight line in joint space between start and goal states), parametrized by the process noise covariance $\mathbf{Q}_c$, generates a kernel which penalizes deviation from the prior in a planning context.
  This clearly corresponds to the smoothness functional $\mathcal{F}\left[\boldsymbol{\theta}(t)\right]$ in the trajectory optimization definition from Eq.~\eqref{eq:trajopti}.

  The likelihood $l ( \boldsymbol { \theta } ; \bar{S} )$ assumes the role of $\mathcal{M}\left[\boldsymbol{\theta}(t)\right]$, with the covariance $\boldsymbol{\Sigma}_{\bar{S}}$ serving as the weighing term $\lambda$.
  Similarly, the collision-free likelihood $ l ( \boldsymbol { \theta } ; \bar{C}  )$, defined in~\cite{gpmp-ijrr} (derived from~\cite{ratliff2009chomp}) as the signed distance of the robot's body from an obstacle, naturally represents the collision avoidance functional $\mathcal{C}\left[\boldsymbol{\theta}(t)\right]$.
  In~\cite{gpmp-ijrr}, likelihoods constraining end-effector position and orientation over support states are also defined.
\section{Experiments}
In this section, we demonstrate how our formulation can be used to optimize constrained trajectories, as well as to avoid constraint-induced singularities, and show how the GP trajectory representation can help make this process more efficient.
We also compare the approach to manipulability maximization presented in this paper to existing control-based formulations, by quantitatively evaluating performance on a constrained planning problem with randomized initial manipulator configurations.
Finally, we present results for a singularity avoidance scenario involving a real Universal Robots UR-10 manipulator available in our laboratory at the University of Toronto, which is able to move throughout a 6-DOF task space that contains many singular and near-singular configurations. All planning computations were performed on a laptop with an Intel i7-8750H CPU running at 2.20 GHz and with 16 GB of RAM.
\subsection{Maximizing Manipulability}\label{subsec:conf_goal}
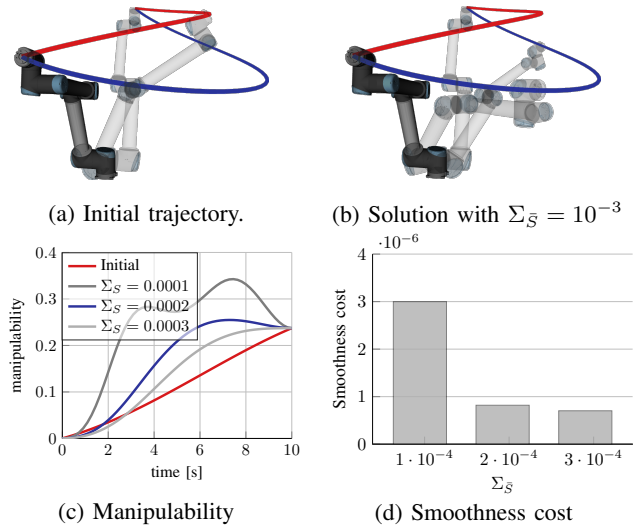
\begin{figure}
  \begin{subfigure}[]{0.49\columnwidth}
    \vspace{3mm}
    \centering    
    \includegraphics[width=0.8\columnwidth]{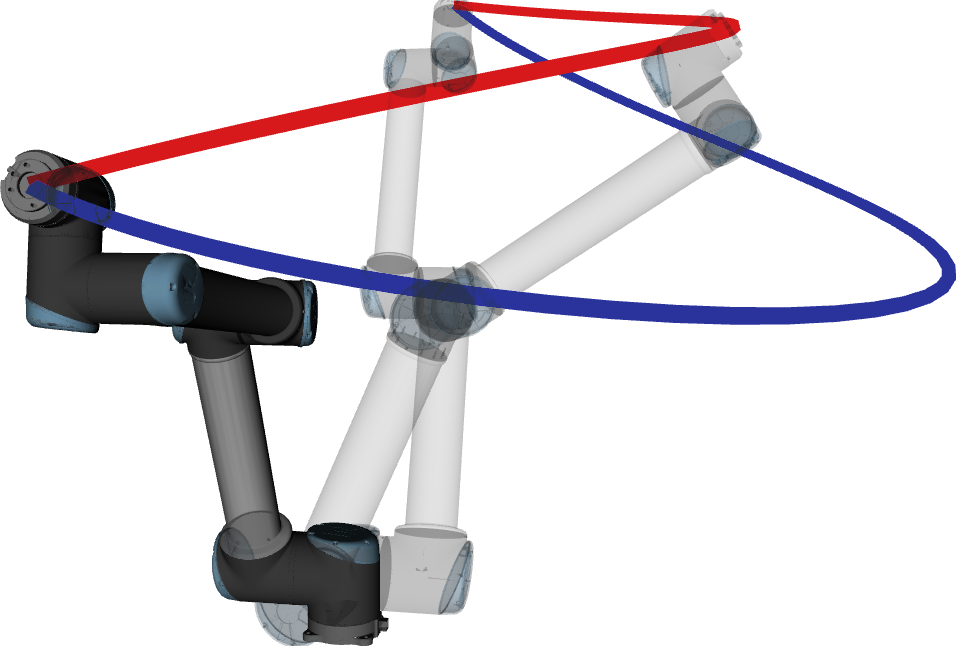}
    \vspace{1mm}
    \caption{Initial trajectory.}
    \label{fig:scenario_1_viz_a}
  \end{subfigure}
  \begin{subfigure}[]{0.49\columnwidth}
    \vspace{3mm}
    \centering
    \includegraphics[width=0.8\columnwidth]{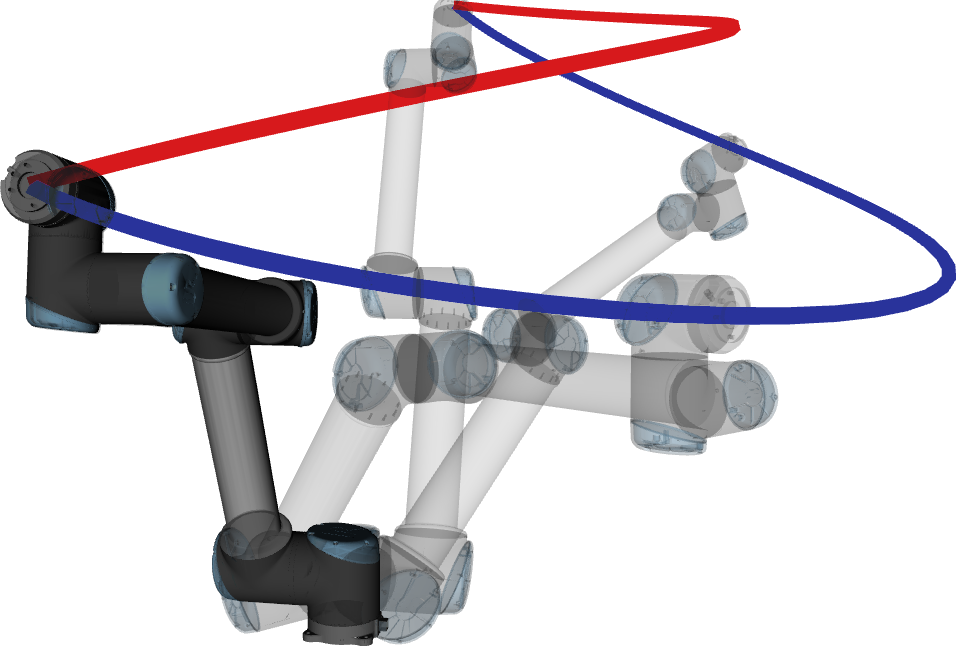}
    \vspace{1mm}
    \caption{Solution with $\Sigma_{\bar{S}} = 10^{-3}$}
    \label{fig:scenario_1_viz_b}
  \end{subfigure}\vspace{0.5mm}
  \begin{subfigure}[]{0.49\columnwidth}
    \vspace{1mm}
    \centering
    \resizebox{\textwidth}{!}{
%
%
\definecolor{mycolor1}{RGB}{215,25,28} 
\definecolor{mycolor3}{RGB}{101,193,63} 
\definecolor{mycolor3}{RGB}{45,51,155} 
\definecolor{mycolor2}{RGB}{125,125,125}
\definecolor{mycolor4}{RGB}{175,175,175}

\begin{tikzpicture}

\begin{axis}[%
width=4.514in,
height=3.662in,
at={(0.757in,0.481in)},
scale only axis,
xmin=0,
xmax=10,
xlabel style={font=\color{white!15!black}},
xlabel={time [s]},
xlabel near ticks,
ymin=0,
ymax=0.4,
ylabel style={font=\color{white!15!black}},
ylabel={manipulability},
axis background/.style={fill=white},
axis x line*=bottom,
axis y line*=left,
scale =0.43,
ylabel near ticks,
xmajorgrids,
ymajorgrids,
legend style={at={(0.59,1
    )}, legend cell align=left, align=left, draw=white!15!black, text opacity =1, fill opacity = 0.6}
]
\addplot [color=mycolor1, line width=1.5pt]
  table[row sep=crcr]{%
0	0.00013596800757758\\
0.24390243902439	0.00357124310327096\\
0.48780487804878	0.0072384650224424\\
0.731707317073171	0.0111327940086275\\
0.975609756097561	0.0152488463953645\\
1.21951219512195	0.019580705095553\\
1.46341463414634	0.0241219313614557\\
1.70731707317073	0.0288655777887884\\
1.95121951219512	0.0338042025353743\\
2.19512195121951	0.0389298847219608\\
2.4390243902439	0.0442342409799749\\
2.68292682926829	0.0497084431082637\\
2.92682926829268	0.0553432367982473\\
3.17073170731707	0.0611289613843315\\
3.41463414634146	0.0670555705740283\\
3.65853658536585	0.073112654109851\\
3.90243902439024	0.0792894603128773\\
4.14634146341463	0.0855749194557224\\
4.39024390243902	0.0919576679107241\\
4.63414634146341	0.0984260730172642\\
4.8780487804878	0.104968258610436\\
5.1219512195122	0.111572131151696\\
5.36585365853659	0.118225406400666\\
5.60975609756098	0.124915636565999\\
5.85365853658537	0.131630237872015\\
6.09756097560976	0.138356518476841\\
6.34146341463415	0.145081706676951\\
6.58536585365854	0.151792979332254\\
6.82926829268293	0.158477490445402\\
7.07317073170732	0.165122399828533\\
7.31707317073171	0.171714901790509\\
7.5609756097561	0.178242253777582\\
7.80487804878049	0.184691804900542\\
8.04878048780488	0.191051024281637\\
8.29268292682927	0.197307529154968\\
8.53658536585366	0.203449112654618\\
8.78048780487805	0.209463771225494\\
9.02439024390244	0.215339731592736\\
9.26829268292683	0.221065477226566\\
9.51219512195122	0.226629774240624\\
9.75609756097561	0.232021696663162\\
10	0.237230651021924\\
};
\addlegendentry{Initial}

\addplot [color=mycolor2, line width=1.5pt]
  table[row sep=crcr]{%
0	0.000135968007584102\\
0.181818181818182	0.000762165578480077\\
0.363636363636364	0.00276805256259427\\
0.545454545454545	0.00655526208517926\\
0.727272727272727	0.0126941538196298\\
0.909090909090909	0.0217820788845185\\
1.09090909090909	0.034203821009534\\
1.27272727272727	0.0500994145909355\\
1.45454545454545	0.0694218791510568\\
1.63636363636364	0.0918080791004398\\
1.81818181818182	0.116573208100102\\
2	0.142728385464922\\
2.18181818181818	0.169083493765759\\
2.36363636363636	0.194407456687681\\
2.54545454545455	0.217555518035071\\
2.72727272727273	0.237604228045097\\
2.90909090909091	0.2540121690967\\
3.09090909090909	0.266569952892235\\
3.27272727272727	0.275309061746616\\
3.45454545454545	0.280538738968312\\
3.63636363636364	0.282800515400168\\
3.81818181818182	0.282802895190182\\
4	0.281288910738252\\
4.18181818181818	0.27894914472792\\
4.36363636363636	0.276400026951133\\
4.54545454545455	0.274168530102936\\
4.72727272727273	0.272677930522939\\
4.90909090909091	0.272249846470402\\
5.09090909090909	0.273118133445779\\
5.27272727272727	0.275426419180328\\
5.45454545454545	0.279229386297073\\
5.63636363636364	0.2844428353315\\
5.81818181818182	0.290834244206265\\
6	0.29812185410599\\
6.18181818181818	0.305969845294224\\
6.36363636363636	0.313999027206299\\
6.54545454545455	0.321780414785897\\
6.72727272727273	0.328863916165714\\
6.90909090909091	0.334833048580709\\
7.09090909090909	0.339315180729491\\
7.27272727272727	0.342001317497818\\
7.45454545454545	0.342661757681229\\
7.63636363636364	0.341152180666781\\
7.81818181818182	0.337433244215642\\
8	0.331578261917769\\
8.18181818181818	0.323770036847339\\
8.36363636363636	0.314311492878768\\
8.54545454545454	0.303623474424299\\
8.72727272727273	0.292196028359486\\
8.90909090909091	0.280568836286626\\
9.09090909090909	0.269304454114116\\
9.27272727272727	0.258960777279207\\
9.45454545454546	0.25008636228486\\
9.63636363636364	0.243202960209623\\
9.81818181818182	0.238782004160221\\
10	0.237230651022405\\
};
\addlegendentry{$\Sigma_S  = 0.0001$}

\addplot [color=mycolor3, line width=1.5pt]
  table[row sep=crcr]{%
0	0.00013596800758735\\
0.181818181818182	0.000390196620343027\\
0.363636363636364	0.00115524742901494\\
0.545454545454545	0.00246218321295516\\
0.727272727272727	0.00436859227936526\\
0.909090909090909	0.0069464750454499\\
1.09090909090909	0.0102546158833461\\
1.27272727272727	0.0143442745278255\\
1.45454545454545	0.0192713033249884\\
1.63636363636364	0.0250753976343594\\
1.81818181818182	0.0317754275346737\\
2	0.0393615885631988\\
2.18181818181818	0.0477932659199089\\
2.36363636363636	0.057003587501689\\
2.54545454545455	0.0668977775106389\\
2.72727272727273	0.0773556040818126\\
2.90909090909091	0.0882654291933093\\
3.09090909090909	0.0995246370742896\\
3.27272727272727	0.111002651286602\\
3.45454545454545	0.122562619547259\\
3.63636363636364	0.134065034961565\\
3.81818181818182	0.145376197515882\\
4	0.156398689208734\\
4.18181818181818	0.167053593768537\\
4.36363636363636	0.177270258350598\\
4.54545454545455	0.186986286434255\\
4.72727272727273	0.196148198287619\\
4.90909090909091	0.204707290926708\\
5.09090909090909	0.212619892863715\\
5.27272727272727	0.219849104683503\\
5.45454545454545	0.22636399702016\\
5.63636363636364	0.232152803902474\\
5.81818181818182	0.237229897968225\\
6	0.24160953123423\\
6.18181818181818	0.245308189012821\\
6.36363636363636	0.248344092908182\\
6.54545454545455	0.250746033492769\\
6.72727272727273	0.252556085357723\\
6.90909090909091	0.253814033082719\\
7.09090909090909	0.25456050510552\\
7.27272727272727	0.254837096445809\\
7.45454545454545	0.254681537423778\\
7.63636363636364	0.254133757430649\\
7.81818181818182	0.253240752367514\\
8	0.252051249397058\\
8.18181818181818	0.250615651596619\\
8.36363636363636	0.248993769093738\\
8.54545454545454	0.2472547224971\\
8.72727272727273	0.245464960835683\\
8.90909090909091	0.243691844885221\\
9.09090909090909	0.242003223086124\\
9.27272727272727	0.240466120944804\\
9.45454545454546	0.239150364750655\\
9.63636363636364	0.238127061863671\\
9.81818181818182	0.237465212803567\\
10	0.237230651022026\\
};
\addlegendentry{$\Sigma_S  = 0.0002$}

\addplot [color=mycolor4, line width=1.5pt]
  table[row sep=crcr]{%
0	0.00013596800758072\\
0.181818181818182	0.000322904831602265\\
0.363636363636364	0.000880187716638655\\
0.545454545454545	0.00181405010495271\\
0.727272727272727	0.00314257130869964\\
0.909090909090909	0.00489145613229794\\
1.09090909090909	0.00708361286524706\\
1.27272727272727	0.00974209532323591\\
1.45454545454545	0.0128945970945282\\
1.63636363636364	0.0165649481605298\\
1.81818181818182	0.0207710468024788\\
2	0.0255214690997603\\
2.18181818181818	0.0308139331070239\\
2.36363636363636	0.0366359821081156\\
2.54545454545455	0.0429632768006157\\
2.72727272727273	0.0497592915974342\\
2.90909090909091	0.0569871657277344\\
3.09090909090909	0.0646094270092665\\
3.27272727272727	0.0725731296095223\\
3.45454545454545	0.0808174584317535\\
3.63636363636364	0.0892747224181346\\
3.81818181818182	0.0978736039729919\\
4	0.106552461336928\\
4.18181818181818	0.115253252098318\\
4.36363636363636	0.123918019770621\\
4.54545454545455	0.132489622020078\\
4.72727272727273	0.140912886238455\\
4.90909090909091	0.149133671025435\\
5.09090909090909	0.157099546392057\\
5.27272727272727	0.164761328827521\\
5.45454545454545	0.172073486498653\\
5.63636363636364	0.179001506101837\\
5.81818181818182	0.185526174627181\\
6	0.191631104067751\\
6.18181818181818	0.197304100961521\\
6.36363636363636	0.202536933396505\\
6.54545454545455	0.207329671490054\\
6.72727272727273	0.211691740422867\\
6.90909090909091	0.215633851523183\\
7.09090909090909	0.219169228584227\\
7.27272727272727	0.222313260341517\\
7.45454545454545	0.225080751630359\\
7.63636363636364	0.227488598493191\\
7.81818181818182	0.229557703019337\\
8	0.231309982344249\\
8.18181818181818	0.232768051299208\\
8.36363636363636	0.233958559809977\\
8.54545454545454	0.234911795221934\\
8.72727272727273	0.235655914032882\\
8.90909090909091	0.236218537894232\\
9.09090909090909	0.236626509539243\\
9.27272727272727	0.236905296546493\\
9.45454545454546	0.237080646612727\\
9.63636363636364	0.237177936059685\\
9.81818181818182	0.237220765235183\\
10	0.237230651021955\\
};
\addlegendentry{$\Sigma_S  = 0.0003$}
\end{axis}
\end{tikzpicture}%
}
    \caption{Manipulability}
    \label{fig:scenario_1_mnp}
  \end{subfigure}
  \begin{subfigure}[]{0.49\columnwidth}
    \centering
    \resizebox{\textwidth}{!}{
%
%
\definecolor{mycolor2}{RGB}{101,193,63}
\definecolor{mycolor1}{RGB}{125,125,125}%
\definecolor{mycolor3}{RGB}{175,175,175}%
\definecolor{mycolor4}{RGB}{215,25,28}
\begin{tikzpicture}

\begin{axis}[%
width=4.514in,
height=3.302in,
at={(0.757in,0.481in)},
scale only axis,
bar shift auto,
xmin=0.5,
xmax=4.5,
xlabel style={font=\color{white!15!black}},
xlabel = {$\Sigma_{\bar{S}}$},
xlabel near ticks,
axis x line*=bottom,
xtick={1.3, 2.6, 3.9},
xticklabels = {$1\cdot 10^{-4}$, $2\cdot 10^{-4}$, $3\cdot 10^{-4}$},
ymin=0,
ymax=4e-06,
ylabel style={font=\color{white!15!black}},
ylabel={Smoothness cost},
ylabel near ticks,
axis y line*=left,
ymajorgrids,
scale = 0.45,
axis background/.style={fill=white},
legend style={legend cell align=left, align=left, draw=white!15!black},
legend image post style={sharp plot, line width=0pt, mark=none}
]
\addplot[ybar, bar width=30, fill=mycolor3, fill opacity=0.8, draw=black, draw opacity = 0.5, area legend] table[row sep=crcr] {%
2.1	2.99850126607073e-06\\
};
\addlegendentry{$\Sigma_{\bar{S}}= 0.0001$}

\addplot[ybar, bar width=30, fill=mycolor3, fill opacity=0.8, draw=black,  draw opacity = 0.5, area legend] table[row sep=crcr] {%
2.5	8.18820994331112e-07\\
};
\addlegendentry{$\Sigma_{\bar{S}}= 0.0002$}

\addplot[ybar, bar width=30, fill=mycolor3, fill opacity=0.8, draw=black,  draw opacity = 0.5, area legend] table[row sep=crcr] {%
2.9	7.02212584731863e-07\\
};
\addlegendentry{$\Sigma_{\bar{S}}= 0.0003$}

\addplot[forget plot, color=white!15!black] table[row sep=crcr] {%
-0.2	0\\
5.2	0\\
};
\legend{}
\end{axis}
\end{tikzpicture}%
}
    \vspace{-6mm}
    \caption{Smoothness cost}
    \label{fig:scenario_1_smt}
  \end{subfigure}
  \caption{Reaching a goal configuration. Top: Visualization, the final state is opaque. Bottom: The manipulability and smoothness costs for different $\Sigma_{\bar{S}}$.}
  \label{fig:simulated_scenarios_a}
  \vspace*{-6mm}
\end{figure}
In the scenario shown in Fig.~\ref{fig:simulated_scenarios_a}, the goal configuration needs to be reached by the manipulator in $T = 10 \text{ s}$.
A simple linear interpolation in joint space between the start and goal configurations is used as an initializing trajectory, with 11 support states, 89 interpolated states, $\mathbf{Q}_c = 10^{5}\,\mathbf{I}$, and with the covariance in Eq.~\eqref{eq:prior2} of the start and goal state priors set to $\boldsymbol{\Sigma}_{\theta} = 10^{-3}\mathbf{I}$.\footnote{Unless specified otherwise, we use these parameters in all experiments.}
Examining the manipulability of the initial trajectory in Fig.~\ref{fig:scenario_1_mnp}, we can see that the starting configuration at $t = 0$ is clearly singular.
After optimizing the joint trajectory to maximize our likelihood function (i.e., Eq.~\ref{eq:mnp_likelihood}) using the method described in Section~\ref{section:trajopt}, the time spent in high manipulability configurations is much greater; the computation (planning) time is $20 \, \text{ms}$.
Figs.~\ref{fig:scenario_1_viz_a} and~\ref{fig:scenario_1_viz_b} illustrate that, because we have only constrained the maximization by fixing the starting and final manipulator configuration and the velocity, the optimized trajectory results in a different end-effector path (blue line) than the initialization (red line).
The covariance parameter $\mathbf{\Sigma}_{\bar{S}}$ serves as a gain parameter, where lower values put greater weight on the manipulability cost term.
As shown in Figs.~\ref{fig:scenario_1_mnp} and \ref{fig:scenario_1_smt}, a greater $\mathbf{\Sigma}_{\bar{S}}$ will reduce the effect of manipulability maximization, while a very low value will suppress overall smoothness, leading to a noisy and infeasible result.
Even though trajectory optimization methods like the one described in Section~\ref{section:trajopt} find only a locally optimal solution, formulating the problem by constraining the trajectory in this manner results in low sensitivity to the exact parametrization when a reasonable initialization is used.

\subsection{Collision Avoidance}
\label{sec:coll}

\begin{figure}
  \begin{subfigure}[]{0.49\columnwidth}
    \centering
    \includegraphics[width=0.7\columnwidth]{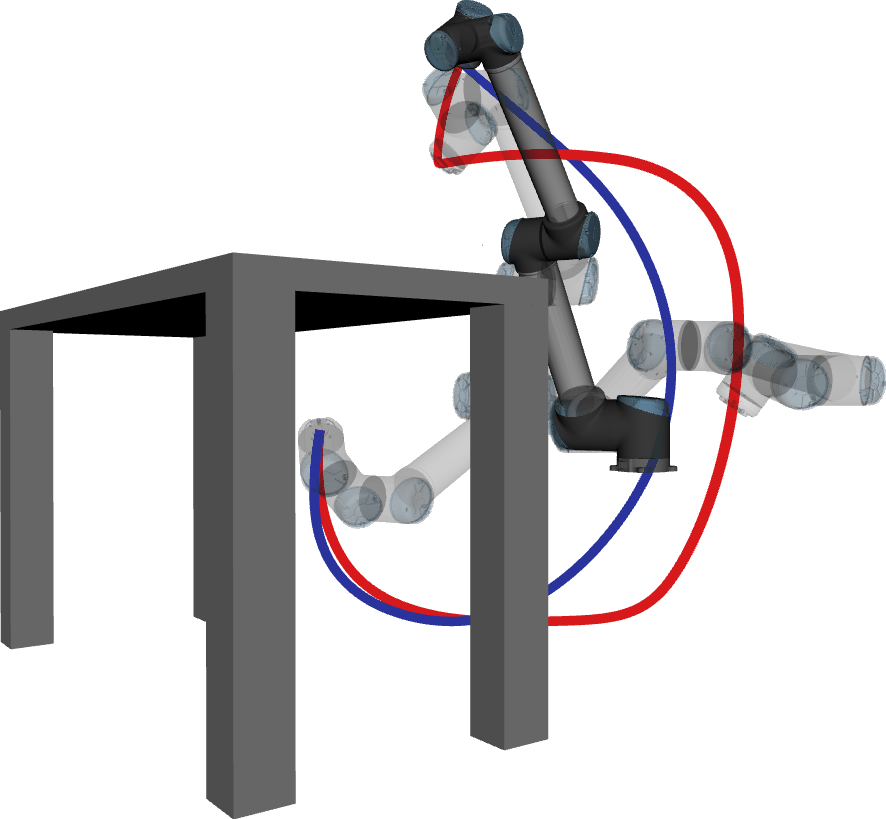}
    \caption{Initial trajectory}
    \label{fig:scenario_obs_viz_a}
  \end{subfigure}
  \begin{subfigure}[]{0.49\columnwidth}
    \centering
    \includegraphics[width=0.7\columnwidth]{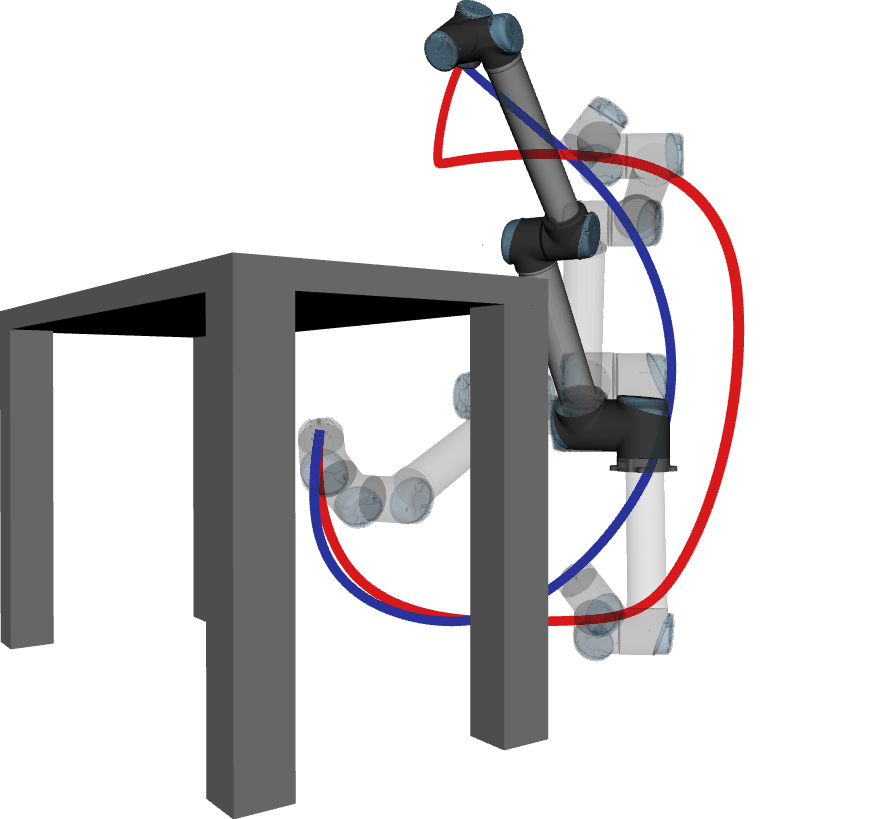}
    \caption{Solution.}\label{fig:scenario_obs_viz_b}
  \end{subfigure}\\
  \begin{subfigure}[]{0.49\columnwidth}
    \vspace{2mm}
    \centering
    \resizebox{\textwidth}{!}{\input{figures/sim_mnp/final/rev/obstacle_mnp_rev.tex}}
    \caption{Manipulability}\label{fig:simulated_scenarios_obs_mnp}
  \end{subfigure}
  \begin{subfigure}[]{0.49\columnwidth}
    \vspace{2mm}
    \centering
    \resizebox{\textwidth}{!}{\input{figures/sim_mnp/final/rev/obstacle_sng_rev.tex}}
    \caption{Smallest singular value}\label{fig:simulated_scenarios_obs_lsv}
  \end{subfigure}
  \caption{Reaching a goal configuration while avoiding an obstacle. Top: Visualization, the final state is opaque. Bottom: The manipulability and smallest singular values with and without the use of interpolated states.}\label{fig:simulated_scenarios_obs}
  \vspace{-6mm}
\end{figure}

Manipulability is often overlooked in tasks that include collision avoidance.
In Fig.~\ref{fig:scenario_obs_viz_a}, we visualize a trajectory generated by a motion planning algorithm that reaches a goal configuration while avoiding collision with a table-shaped obstacle in the workspace.
The manipulability index and smallest singular value plots in Figs.~\ref{fig:simulated_scenarios_obs_mnp} and \ref{fig:simulated_scenarios_obs_lsv} show that this trajectory contains singularities induced by collision avoidance.
If a change in task space constraints were to happen during trajectory execution, and if this change required a rapid response (e.g., if another obstacle appeared or a collaborator pushed the end-effector in a certain direction), the poor conditioning of the Jacobian matrix would cause violent joint movements.
To give a numerical example, had this happened at the $6.5 \text{ s}$ mark, generating an end-effector velocity of 1 cm/s in the $x$ direction in task space by computing the pseudo-inverse of the kinematic relationship in Eq.~\eqref{eq:jacobian} would have resulted in joint velocities exceeding $150~\text{rad/s}$.

This possibility is prevented by optimizing the initial trajectory for (locally) maximum manipulability.
 In addition to the start and goal states, we set the state prior at the fifth support state ($4.5 \text{ s}$) to that of the initial trajectory, with a covariance of $\Sigma_\theta = 10^3\,\mathbf{I}$.
Since the initial trajectory successfully avoids collision, this improves the default straight line prior; multiple states can be fixed in this manner.
To maintain distance from the obstacle, we add collision avoidance factors as described in~\cite{gpmp-ijrr} to each support state, with the parameters $\Sigma_{obs} = 10^2\,\mathbf{I} $ and $\epsilon = 0.3$.
 Due to the local nature of the trajectory optimization method, the choices for $\Sigma_{obs}$ and $\Sigma_{\bar{S}}$ determine the trade off between collision avoidance and manipulability maximization, and so this selection needs to be made carefully.
The optimization is carried out within $5 \, \text{ms}$ and $50 \, \text{ms}$ without and with interpolated states, respectively.
  Fig.~\ref{fig:scenario_obs_viz_b} shows that collisions, as well as the singularities caused by collision avoidance, are completely prevented.

\subsection{Reaching Task}\label{subsec:reaching}

Reaching a Cartesian goal point with the end-effector is a very common manipulation task, often performed in situations where unexpected changes in the task are possible and where maintaining high manipulability is therefore very important (e.g., in kinesthetic teaching).
As mentioned in Section~\ref{section:related_work}, most manipulability maximization methods use kinematic control or global optimization to follow a path initialized in Cartesian space.
The local nature of our approach allows for fast computation, comparable to~\cite{zhang2016qp, dufour2017integrating}, when used for planning a trajectory.

To further illustrate the benefits of our technique, we compare with the QP formulation in~\cite{zhang2016qp} and the established singularity avoidance approach in~\cite{chiaverini1997singularity}.
We plan a trajectory to reach a desired end-effector position from 50 random (feasible) initial configurations.
Here, a joint-space trajectory (position and velocity profile) must be generated, with a time step of $T = 0.02$, that maximizes the average manipulability and maintains joint velocities below $\frac{\pi}{3}~\text{rad/s}$.
We initialize the problem with 50 support states, parametrizing the optimization with $\mathbf{Q}_c = 10^{6}\,\mathbf{I}$, $\Sigma_{\bar{S}} = 0.0013$ and $\boldsymbol{\Sigma}_{\theta} = 10^{-3}\,\mathbf{I}$.
To identify the final configuration used to initialize the trajectory, we iterate over 20 possible inverse kinematics solutions generated by the \texttt{quadprog} function in MATLAB.
We avoid initializations that inevitably pass through a singularity by using the fast interpolation property to choose an initialization with the greatest minimum manipulability index.
\begin{table*}
  \begin{center}
    \caption{Trajectory generation performance comparison for reaching task.}
    \label{table:comparison_1}
    \begin{tabular}{lccccccccc}  
      \toprule
      & \multicolumn{3}{c}{Manipulability} & \multicolumn{2}{c}{Velocities \text{[rad/s]}} & \multicolumn{3}{c}{Solve Time \text{[s]}} & \\
      \cmidrule(r){2-4}  \cmidrule(r){5-6} \cmidrule(r){7-9}
      & Avg. & Min & Max & Max & Avg & Total & Opt. & Init. & Solved \\
      \midrule
      \footnotesize{This paper - No Intp.}                           &0.1316 &0.0460 &0.1984 &0.2830 &0.2009  & \textbf{0.2560} & 0.0213 & 0.2347 & 50/50\\
      \footnotesize{This paper - Intp.}                              &\textbf{0.1422} &0.0450 &\textbf{0.2206} &0.3529 &0.2425  &0.3158 & 0.0870 & 0.2288 & 50/50\\
      \footnotesize{Method in~\cite{zhang2016qp}}                    &0.1385 &0.0481 &0.2077 &0.3178 &0.1173 &1.5064 & 1.5064 & 0 & 46/50\\
      \footnotesize{Method in~\cite{chiaverini1997singularity}}      &0.0748 &0.0286 &0.1193 &0.1310 &0.0482 &0.8155 & 0.8155 & 0 & 50/50\\
      \bottomrule
    \end{tabular}
  \end{center}
  \vspace{-2mm}
\end{table*}
\begin{figure}
  \centering
  \begin{subfigure}[]{0.9\columnwidth}
    \centering
    \resizebox{\textwidth}{!}{
%
%
\definecolor{mycolor3}{RGB}{204,235,197}%
\definecolor{mycolor4}{RGB}{222,203,228}%
\definecolor{mycolor5}{RGB}{254,217,166}%
\definecolor{mycolor2}{RGB}{125,125,125}%
\definecolor{mycolor1}{RGB}{200,200,200}%
\begin{tikzpicture}

\begin{axis}[%
width=4.514in,
height=1.3in,
at={(0.757in,0.481in)},
scale only axis,
bar shift auto,
xmin=0.507692307692308,
xmax=5.49230769230769,
xtick={1, 2, 3, 4, 5},
xticklabels={1,5,10,20,50},
ymin=0,
ymax=0.2,
ytick={0, 0.1, 0.2},
ymajorgrids,
ylabel={Avg. Manipulability},
ylabel near ticks,
xlabel = {Initializations},
xlabel near ticks,
scale = 0.9,
axis background/.style={fill=white},
legend style={at={(0.23,0.95)}, legend cell align=left, align=left, draw=white!15!black}
]
\addplot[ybar, bar width=20, fill=mycolor1, fill opacity = 0.8, draw=black, draw opacity = 0.5, area legend] table[row sep=crcr] {%
1	0.0848\\
2	0.1097\\
3	0.1251\\
4   0.1305\\
5   0.1372\\
};
\addplot[forget plot, color=white!15!black] table[row sep=crcr] {%
0.507692307692308	0\\
4.49230769230769	0\\
};
\addlegendentry{No Intp.}

\addplot[ybar, bar width=20, fill=mycolor2, fill opacity = 0.8, draw=black, draw opacity = 0.5, area legend] table[row sep=crcr] {%
1	0.0989\\
2	0.1167\\
3	0.1296\\
4   0.1412\\
5   0.1465\\
};
\addplot[forget plot, color=white!15!black] table[row sep=crcr] {%
0.507692307692308	0\\
4.49230769230769	0\\
};
\addlegendentry{Intp.}

\end{axis}
\end{tikzpicture}%
}
  \end{subfigure}
  \begin{subfigure}[]{0.9\columnwidth}
    \centering
    \resizebox{\textwidth}{!}{
%
%
\definecolor{mycolor3}{RGB}{204,235,197}%
\definecolor{mycolor4}{RGB}{222,203,228}%
\definecolor{mycolor5}{RGB}{254,217,166}%
\definecolor{mycolor2}{RGB}{125,125,125}%
\definecolor{mycolor1}{RGB}{200,200,200}%
\begin{tikzpicture}

\begin{axis}[%
width=4.514in,
height=1.3in,
at={(0.757in,0.481in)},
scale only axis,
bar shift auto,
xmin=0.507692307692308,
xmax=5.49230769230769,
xtick={1, 2, 3, 4, 5},
xticklabels={1,5,10,20,50},
ymin=0,
ymax=0.8,
ymajorgrids,
ylabel={Time [s]},
ylabel near ticks,
xlabel = {Initializations},
xlabel near ticks,
scale = 0.9,
axis background/.style={fill=white},
legend style={legend cell align=left, align=left, draw=white!15!black}
]
\addplot[ybar, bar width=20, fill=mycolor1, fill opacity = 0.8, draw=black, draw opacity = 0.5, area legend] table[row sep=crcr] {%
1	0.0346\\
2	0.0844\\
3	0.1434\\
4	0.2658\\
5   0.5997\\
};
\addplot[forget plot, color=white!15!black] table[row sep=crcr] {%
0.507692307692308	0\\
4.49230769230769	0\\
};
\addlegendentry{gpmp}

\addplot[ybar, bar width=20, fill=mycolor2, fill opacity = 0.8, draw=black, draw opacity = 0.5, area legend] table[row sep=crcr] {%
1	0.0977\\
2	0.1543\\
3	0.2093\\
4	0.3331\\
5   0.6656\\
};
\addplot[forget plot, color=white!15!black] table[row sep=crcr] {%
0.507692307692308	0\\
4.49230769230769	0\\
};
\addlegendentry{ch}


\legend{}
\end{axis}
\end{tikzpicture}%
}
  \end{subfigure}
  \caption{Performance of our method for the task described in Section \ref{subsec:reaching} with respect to the number of initial trajectories computed. Top: Average manipulability during task execution. Bottom: Solve time.}
  \label{fig:comparison_1}
  \vspace{-4mm}
\end{figure}
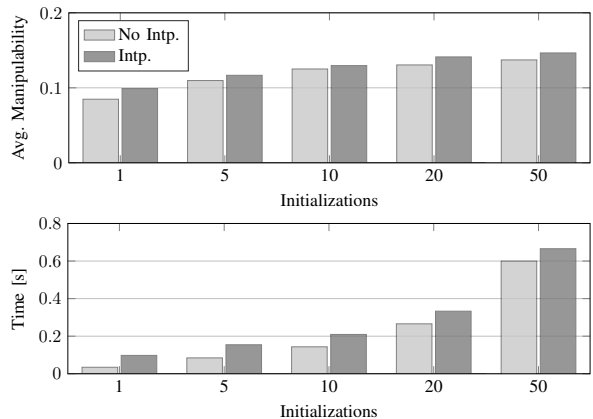

In Table~\ref{table:comparison_1}, we compare the results produced by these three algorithms by examining the (average) minimum, maximum, and mean trajectory manipulability scores, as well as the computation times and the joint velocities involved.
It is clear that, even when searching over a large number of possible initializations, our method achieves dramatically lower computation times.
Consequently, it reaches the highest average and maximum manipulabilities, with the average lowest manipulability index value comparable to~\cite{zhang2016qp}.
The computation time and success rate for~\cite{zhang2016qp} matches the comparison in~\cite{jin2017manipulability}, where it is concluded that the volatility of the manipulability gradient has a significant effect on convergence.
The method in~\cite{chiaverini1997singularity} primarily focusses on avoiding singularities, and also reaches the lowest manipulability index values.
However, it converges faster than~\cite{zhang2016qp} and it always finds a solution.

Initializing over multiple final states (IK solutions) helps avoid singular configurations that would inevitably need to be traversed in joint space, given a bad initialization.
Consequently, the smoothness cost is centered around a trajectory with the minimum number of singularities, which we can then easily optimize as described in Section~\ref{subsec:conf_goal}.
It is worth noting that most of the computation time for our method is taken up by finding the best initialization; this averages around $0.22~\text{s}$ for a 6-DOF manipulator.
We posit that this could be substantially reduced by employing task- or robot-specific heuristics\footnote{For example, avoiding initializations that pass through the elbow singularity, which is trivial to do.}, or fast kinematics solvers like TRAC-IK \cite{beeson2015trac}.
Our hypothesis is supported by the results in Fig.~\ref{fig:comparison_1}, where the average manipulability value clearly rises with the number of precomputed solutions.
There is a diminishing return, however, as increasing the number of solutions past 20 has little effect on the average value, with computation time growing proportionally.
\vspace{-0.5mm}

\subsection{Experiments on a Real Robot}
\begin{figure*}
  \centering
  \begin{subfigure}[]{0.85\columnwidth}
    \centering
    \includegraphics[width=1\columnwidth]{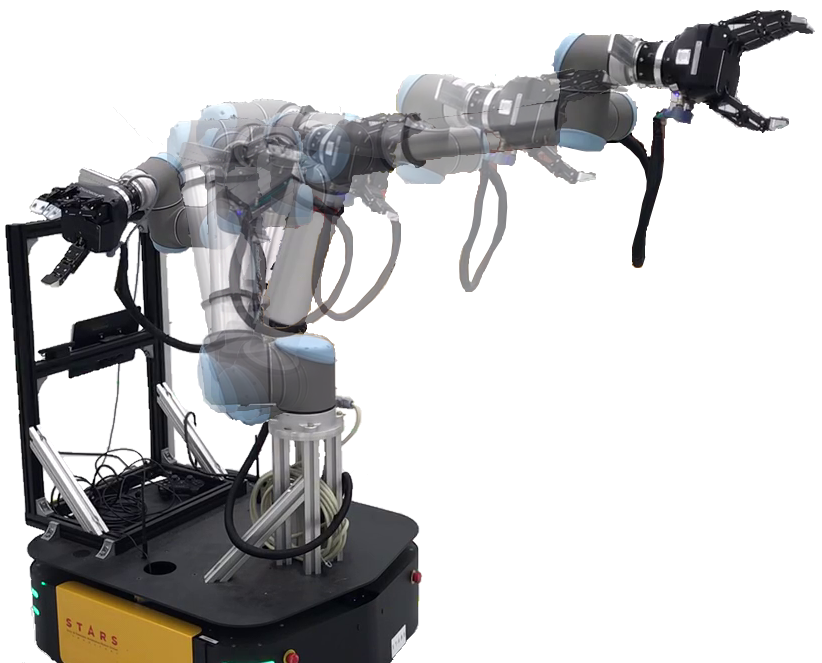}
  \end{subfigure}\hspace{2cm}
  \begin{subfigure}[]{0.85\columnwidth}
    \centering
    \includegraphics[width=1\columnwidth]{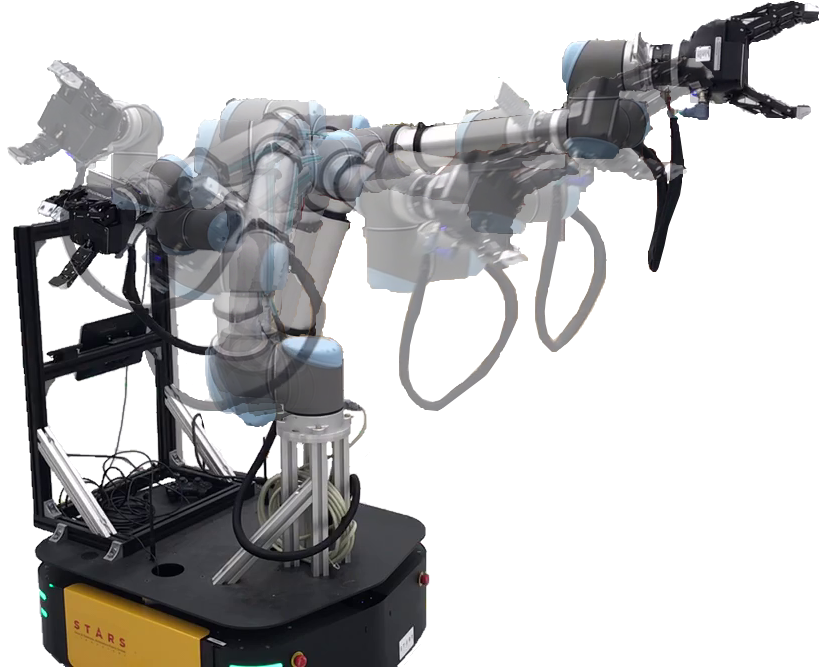}
  \end{subfigure}
  \caption{Experiment on real manipulator. The left image shows a trajectory generated using~\cite{zhang2016qp}, while the image on the right shows a trajectory generated by our method. }
  \label{fig:real}
  \vspace{-2mm}
\end{figure*}
\begin{figure}
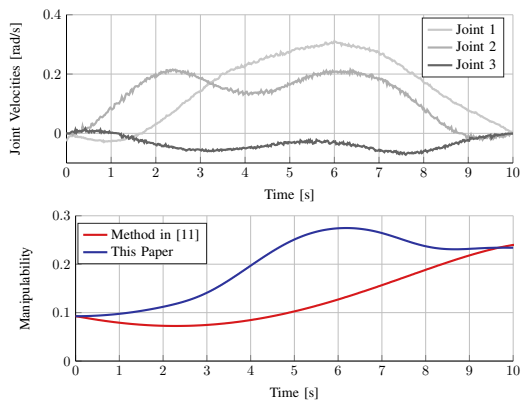

  \centering
  \begin{subfigure}[]{0.8\columnwidth}
    \centering
    \resizebox{\textwidth}{!}{\input{figures/sim_mnp/final/real_vel.tex}}
  \end{subfigure}\\
  \begin{subfigure}[]{0.8\columnwidth}
    \centering
    \resizebox{\textwidth}{!}{\input{figures/sim_mnp/final/real_mnp.tex}}
  \end{subfigure}
  \caption{Top: Velocities of the three largest UR-10 joints using our method. Bottom: Manipulability values compared to~\cite{zhang2016qp}.}
  \label{fig:real_mnp}
  \vspace{-2mm}
\end{figure}

Lastly, we show that the trajectories generated using our method are smooth and can be executed on a real UR-10 manipulator.
The goal is again to generate a maximum manipulability trajectory reaching a final state which is constrained by end-effector position.
We pick a starting configuration that situates the manipulator to one side, as shown in Fig.~\ref{fig:real}. 
In Fig.~\ref{fig:real_mnp} we can see that the velocities generated using our method remain smooth and relatively low, with some noise caused by dynamic effects.
As expected, the method in~\cite{zhang2016qp} follows the shortest Cartesian path towards the goal position, reaching a similar final configuration to that of our method.
However, our method maintains a higher manipulability value throughout the motion.

\section{Conclusion and Future Work}\label{section:future_work_conclusion}

We have presented a novel trajectory planning and replanning method that both maximizes the overall manipulability along a trajectory and inherently avoids singularities.
Our work shows that maximizing manipulability is highly useful for tasks that must be carried out in uncontrolled environments, or when additional constraints are present.
Unlike tracking methods that are commonly used for trajectory planning, our method searches for solutions in joint space, allowing for the efficient use of prior information about the task at hand.
This results in lower computation times and greater overall performance for many tasks.
As an interesting avenue for future work, we believe that our method can be extended to end-effector path tracking tasks by projecting the gradient information into the appropriate null space.

\balance
\bibliographystyle{IEEEcaps}

\end{document}